\documentclass[11pt]{article}

\usepackage[preprint]{acl}

\usepackage{times}
\usepackage{latexsym}
\usepackage{amsmath}
\usepackage{enumitem}
\usepackage{tabularx}
\usepackage{array}
\usepackage{booktabs}

\usepackage[T1]{fontenc}
\usepackage[utf8]{inputenc}

\usepackage{microtype}

\usepackage{inconsolata}

\usepackage{graphicx}

\title{A Computational Approach to Measuring Semantic Change in Sanskrit Literature}

\author{Tanay Agrawal \\
The Harker School, San Jose CA, 95129\\
\texttt{27tanaya@students.harker.org}}

\begin{document}
\maketitle
\begin{abstract}
Diachronic word embeddings have become the modern standard for tracking semantic change, yet they have been largely validated on modern, high-resource, and well segmented languages. This paper tests whether the paradigm transfers to Sanskrit, an ancient, low-resource language whose phonological fusion (\textit{sandhi}), morphological inflection, compounding, and polysemy pose a unique challenge. I assemble a 2.7M-token corpus spanning four canonical periods, recover word boundaries with a neural byte-level sandhi splitter and lemmatizer, and train per-period embeddings across configurations. To evaluate the system, I curate a validation set from historical scholarship and test recovery directionally with anchor displacement. Of 21 testable shifts, 19 move in the philologically attested direction (sign test, $p\approx1.1\times10^{-4}$). I further show which configuration the language forces and comment on opportunities for improvement.
\end{abstract}

\section{Introduction}

Sanskrit is attested across roughly three millennia, and its lexicon carries some of the best-documented semantic changes in historical linguistics. \textit{Asura} shifts from `lord, mighty one' in the early \d{R}g-Veda, where it functions as an epithet of the greatest gods, to `demon', mirroring in reverse the development of its Iranian cognate \textit{ahura} \citep{kamboj}. \textit{Yoga} travels from `the act of yoking' to `spiritual discipline', before later developing into a modern-day fitness practice. These trajectories have been painstakingly chronicled through generations of philological research, which raises the question of whether shifts in Sanskrit word meanings are detectable computationally.

Diachronic word embeddings are the standard tool for this task: train embeddings per time period, relate the resulting spaces, and measure how each word's position changes \citep{hamilton2016}. However, the framework has been validated primarily on modern, high-resource languages with mature computational resources. Applying these methods to Sanskrit presents challenges that have yet to be systematically addressed \cite{sandhan-etal-2023-sanskritshala}. Sandhi refers to the phonological rules that dictate the fusion of adjacent words \citep{whitney1879sanskrit}, obscuring token boundaries: 
\begin{equation}
\textit{iti} \; \text{`thus'} + \textit{uv\=aca} \; \text{`said'}
\rightarrow
\textit{ityuv\=aca}
\end{equation}
Compounding further merges multiple words into long, often unique, lexical units, as in the following example from Pata\~{n}jali's \textit{Yoga S\=utras} (3.46):
\begin{equation}
\textit{graha\d{n}asvar\=up\=asmit\=anvay\=arthavattvasa\d{m}yam\=ad
}
\end{equation}
Moreover, rich inflection scatters one lexeme over dozens of variants, reducing the frequency of any individual form and increasing sparsity. Although major efforts have digitally preserved Sanskrit texts, available corpora remain relatively sparse by the standards of modern NLP, especially for certain time periods. Therefore, modeling Sanskrit semantic change distributionally will help strengthen the broader application of computational approaches to historical linguistics.

Existing work provides the individual pieces of this problem: philology has recorded Sanskrit semantic change, NLP has developed methods for measuring shifts, and Sanskrit NLP has produced the resources and tools needed for computational analysis. This work brings these components together to answer the following questions:

\begin{enumerate}
  \item Do the embeddings recover semantic shifts testified by Indological scholarship?
  \item Which preprocessing and architecture choices does the language force, and why? 
\end{enumerate}

I build a four-period diachronic corpus ($\sim$2.7M tokens), restore word boundaries with a neural sandhi splitter and lemmatizer, and train per-period embeddings against different preprocessing and architecture configurations. Evaluation is performed through directional tests over a validation set curated from culturally grounded semantic shifts. Consequently, this paper has three main contributions:

\begin{enumerate}\setlength{\itemsep}{0pt}
  \item A diachronic embedding framework for Sanskrit spanning four canonical periods, including a released corpus, preprocessing pipeline, modeling choices, and validation set of documented shifts.
  
  \item A controlled evaluation of Sanskrit semantic change detection, using anchor-displacement tests, matched controls, quantitative metrics, and qualitative studies of semantic trajectories.

  \item An analysis of model limitations, including metaphorical polysemy (\textit{go}), non-monotonic trajectories (\textit{p\=ada}), cosine-space deflation, and frequency-related artifacts.
\end{enumerate}

\section{Related Work}

The study of semantic change has a long intellectual history, extending back to ancient thinkers such as the Neoplatonic philosopher Proclus, who analyzed and classified semantic change \cite{kamboj}. Over time, these observations were formalized into taxonomies of semantic change, such as restriction, expansion, and amelioration \citep{breal1897, bloomfield1933}. For Sanskrit specifically, the most comprehensive treatment is \citet{kamboj}, a book-length study of semantic
change in Sanskrit that catalogs individual shifts with textual attestations spanning the \d{R}g-Veda through classical literature, organizing them by cause (religious, social, cultural) and by type. Individual word histories have also received dedicated studies, e.g.\ \citet{riceGunna} on \textit{gu\d{n}a}, \citet{paoloTejas} on \textit{tejas}, and \citet{lubinVrata} on \textit{vrata}. This literature supplies independently documented shifts, with sources and senses, fixed before any model is trained; it acts here as ground truth for evaluating recovery.

Distributional approaches to semantic change predate neural embeddings
\citep{gulordava-baroni-2011-distributional}, but the field consolidated
around per-period embedding spaces compared over time
\citep{kim-etal-2014-temporal,kulkarni2015statistically}. \citet{hamilton2016} established the now-standard framework of slicing historical corpora into discrete time periods, training separate word embeddings for each period, and aligning the resulting vector spaces with orthogonal Procrustes. The framework was validated on English, German, French, and Chinese, while a companion paper \citep{hamilton2016b} compared global and local-neighborhood measures of semantic change. Surveys by \citet{kutuzov-etal-2018-diachronic} and \citet{tahmasebi2021survey} review major methods and challenges in computational semantic change detection. SemEval-2020 Task~1 \citep{schlechtweg-etal-2020-semeval} standardized evaluation across English, German, Swedish, and Latin. Closest to the setting of this paper is work in Ancient Greek. \citet{rodda2017panta} apply \emph{lemmatized} distributional models to explore semantic change, while \citet{perrone-etal-2019-gasc} introduce a genre-aware Bayesian model to address the conflation of polysemy and semantic change in sparse historical corpora. Sanskrit, however, poses further unique challenges, including sandhi fusion, compounding, and heavy inflection, that must be treated accordingly.

Computational Sanskrit has developed a substantial infrastructure designed for various processing tasks \citep{pradeep202}: the Digital Corpus of Sanskrit with morphosyntactic annotation \citep{dcs}, the lexicon-driven Sanskrit Heritage segmenter \citep{goyal2016design}, and a line of neural word segmentation work, from character-level recurrent and convolutional neural models \citep{hellwig-nehrdich-2018-sanskrit} to dual-decoder sequence models \citep{aralikatte-etal-2018-sanskrit} to the byte-level multitask ByT5-Sanskrit \citep{nehrdich2024modelneedbyt5sanskritunified} used in this study. \citet{sandhan-etal-2023-evaluating} evaluate Sanskrit embeddings on synchronic intrinsic tasks, while more recent work has studied diachronic linguistic evolution (but not semantic drift) in Sanskrit using weakly supervised neural-symbolic methods \citep{hariharan2025}. This paper bridges philological scholarship, diachronic word embeddings, and Sanskrit NLP to evaluate recovery of documented shifts.

\section{Methodology}
\subsection{Data}

I partition the Sanskrit textual tradition into four bins following standard periodization (Table~\ref{tab:corpus}): 
\begin{itemize}[nosep]
  \item Vedic: ritual hymn and liturgy
  \item Upani\d{s}adic: speculative philosophical prose
  \item Epic: heroic, \textit{Mah\=abh\=arata} and \textit{R\=am\=aya\d{n}a}
  \item S\=utra/\'S\=astra: systematic treatises
\end{itemize}
I use literary strata rather than century windows as individual works cannot be dated precisely. Texts are downloaded as Plain Transformation .txt files from GRETIL (\href{https://gretil.sub.uni-goettingen.de/}{Göttingen Register of Electronic Texts in Indian Languages}).

\begin{table}[t]
\centering\small
\begin{tabular}{lrrr}
\toprule
Period & Date & Tokens & Vocab \\
\midrule
Vedic & 1500 -- 600 BCE & 417K  & 10{,}783 \\
Upani\d{s}adic & 700 -- 200 BCE & 228K  & 5{,}493 \\
Epic & 400 BCE -- 400 CE & 1.78M & 30{,}103 \\
S\=utra/\'S\=astra & 200 BCE -- 500 CE & 290K & 8{,}591 \\
\bottomrule
\end{tabular}
\caption{Diachronic corpus (sandhi-split variant). Vocabulary counts types with frequency $\geq 5$.}
\label{tab:corpus}
\end{table}

One property of the corpus shapes interpretation: it is imbalanced, a limitation not uncommon in low-resource settings; the Epic bin is approximately $7.8\times$ the size of the Upani\d{s}adic bin. Because bin size affects the geometry of an embedding space independently of meaning, change measures that contrast structure within periods are favored over raw geometry comparisons across them (\S\ref{sec:change}). Additionally, I verify that the recovery result reflects meaning rather than word frequency, which stems from the corpus size (\S\ref{sec:rq2}).

\subsection{Preprocessing pipeline}
\label{sec:prep}

Distributional models assume the input is a sequence of word tokens. However, Sanskrit obscures word boundaries through sandhi, compounding, and heavy inflection. One lexeme dispersed across many fused surface forms can destroy the co-occurrence signal that embeddings depend on.

Each text therefore passes through a fixed pipeline: (1) removal of editorial metadata, verse markers, comments, and digital detritus via regular expressions and manual checks; (2) transliteration to SLP1, a reversible ASCII encoding designed for Sanskrit NLP tasks \cite{N_J_2025}, with Unicode NFC normalization; (3) segmentation into $\leq$350-character chunks at whitespace boundaries to fit the context window of the neural analyzer; (4) sandhi splitting, including a second pass over residual long compounds; and (5) lemmatization, both with the ByT5-Sanskrit model \citep{nehrdich2024modelneedbyt5sanskritunified}.\footnote{Accessed via this Python package: \url{https://pypi.org/project/dharmamitra-sanskrit-grammar/}.} Neural segmentation was used over rule-based systems to better handle out-of-vocabulary words and noisy data. 

Three corpora are retained from this pipeline to evaluate preprocessing as an experimental variable for this task, namely \textbf{raw} (after
step~3), \textbf{sandhi-split} (after step~4), and \textbf{lemmatized} (after step~5). The effect of segmentation alone is large, raising the vocabulary trackable across all four periods from 801 to 1,345 words, a $+68\%$ increase (\S\ref{sec:rq2}).

\subsection{Per-period embeddings}
\label{sec:emb}
I train independent embeddings per period and vary exactly one factor at a time. Every cell of the grid shares one configuration (Appendix~\ref{app:hyper}), so that any difference traces to the factor under study. The grid crosses the three preprocessing variants with two architectures: \textbf{word2vec} skip-gram \cite{mikolov-etal-2013-distributed} and \textbf{FastText} \citep{bojanowski2017}, which represents a word as the sum of its character $n$-grams; for FastText, $n$-gram ranges of $[3,6]$ and $[6,12]$ are tested.

\subsection{Quantifying change}
\label{sec:change}

Quantitative analysis is restricted to the 1{,}345 words shared by all four periods, with two change statistics computed per word. These metrics serve as a reliability criterion for comparing configurations (\S\ref{sec:rq2}), but they could be used to uncover candidate shifts for philological review.

\textbf{Second-order similarity}: since different periods' spaces have unrelated coordinate frames, a word is compared not by its raw vector but by what it is similar to \citep{hamilton2016b}. For a target word and a period pair, a reference set is first built: the union of the target's top-25 nearest neighbors in each of the two periods, capturing both the company the word leaves and the company it joins. The word's similarity profile is then computed per-period as the vector of its cosine similarities to each reference word. The final change score is one minus the cosine between the two profiles. This method does not require cross-period alignment as similarities are measured within a single period.

\textbf{Procrustes-aligned cosine distance}: this approach accounts for the coordinate mismatch, allowing direct vector comparison. Each period's embedding space is aligned to the Vedic space using orthogonal Procrustes \citep{hamilton2016}, with the rotation estimated from stable, high-frequency words across all periods. After alignment, the change score is simply the cosine distance between the two vectors. However, no rotation can perfectly reconcile the  independently trained spaces, so this measure inevitably carries alignment noise.

% \textbf{Nearest-neighbor Jaccard distance}: calculated as one minus the overlap between the target word's top-$k$ neighbor lists in the two periods. Neighborhood membership is binary and graded similarity values are discarded, so on corpora of this size, the top-$k$ lists fluctuate even for stable words and the metric saturates at its maximum.

\subsection{Evaluating recovery}
\label{sec:recovery}
To quantitatively measure recovery, I pre-register the prediction for each shift and test it directionally with anchor vectors, following \citet{hamilton2016}. 

I first curate a validation set with 21 documented shifts and one stable control from \citet{kamboj}: candidate words were selected from the literature and filtered by corpus frequency before any embedding analysis. Each entry's source and target senses are recorded, along with the change type, and the period transition in which the shift is documented. The full set appears in Appendix~\ref{app:valset}.

For each shift, small anchor sets denoting the old and the new sense are then specified. For \textit{gu\d{n}a}, for example, the old-sense anchors are related to string and binding (\textit{tantu} `thread', \textit{rajju} `cord', \textit{jy\=a} `bowstring'), while the new-sense anchors revolve around the `quality, virtue' meaning (\textit{svabh\={a}va} `inherent nature', \textit{lak\d{s}a\d{n}a} `defining attribute'). Let $c_{\text{old}}(t)$ and $c_{\text{new}}(t)$ be the cosine similarity between the target word and the mean old-sense and new-sense anchor vector in period $t$. A word is considered to move in its historically attested direction if
\begin{equation}
\Delta = \bigl[c_{\text{new}}-c_{\text{old}}\bigr]_{\text{end}} - \bigl[c_{\text{new}}-c_{\text{old}}\bigr]_{\text{start}} > 0
\end{equation}
across its transition from period \texttt{start} to \texttt{end}. This two-sided contrast is important because absolute cosine similarities often decrease as embedding spaces grow more dispersed. Measuring movement relative to both the old and new senses isolates semantic change from changes in embedding-space geometry.

Each lexical item yields a directional prediction (either $\Delta>0 \text{ or } \Delta \leq 0$). Statistical significance is therefore assessed (over the collection of documented shifts) using a one-sided binomial sign test under the null hypothesis that each word is equally likely to move in either direction (p = 0.5). 

Finally, at each methodological step, lexical meanings were cross-checked against external Sanskrit dictionaries.\footnote{Semantic interpretation and verification were performed using the \href{https://www.sanskrit-lexicon.uni-koeln.de/}{Cologne Digital Sanskrit Dictionaries}  (version 2.10.72) interface for Monier-Williams' Sanskrit-English Dictionary, the \href{https://www.wisdomlib.org/sanskrit}{Wisdom Library}, and \href{https://sanskritdictionary.com/}{Sanskrit Dictionary}.}

\section{Results}
\subsection{Recovery of known shifts}
\label{sec:result}

Table~\ref{tab:anchor} gives the directional test for every documented shift. 19 of 21 testable shifts move in the attested direction; per the one-sided sign test, $p\approx0.00011$. 

As a control, I apply the anchor set for \textit{asura} to \textit{deva}, drawing from evidence that shows both \textit{deva} and \textit{asura} originally described powerful beings and gods \cite{kamboj}. The pejoration of Proto-Indo-Iranian \textit{*dayw\'{a}s} occurred only in the Iranian cognate (Avestan \textit{da\={e}va} `demon'), whereas Sanskrit \textit{deva} retained its meaning `god' \cite{Witzel2001Autochthonous}. Similarly, Sanskrit \textit{asura} pejorated toward demonic, while Avestan \textit{ahura} retained a positive meaning \cite{bhargavaAsura}. Table~\ref{tab:asura} illustrates that the embeddings recover this philological trajectory. Furthermore, in the Vedic bin deva and asura are near-equidistant from the divine and demonic anchors (0.65/0.67 and 0.64/0.65), which could signify the dual usage in mighty beings and the shift from early Rigveda to end of Vedic period. Across the transition they diverge in opposite directions: asura moves demonward ($\Delta=+0.199$) while deva consolidates toward the divine anchors ($\Delta=-0.159$).

\begin{table}[t]
\centering\small
\begin{tabular}{@{}lr@{}}
\toprule
Word & ${\Delta}$ \\
\midrule
\textit{hari}        & $+0.677$ \\
\textit{arka}        & $+0.639$ \\
\textit{brahman}     & $+0.387$ \\
\textit{yoga}        & $+0.357$ \\
\textit{k\d{s}atra}  & $+0.344$ \\
\textit{uttama}      & $+0.322$ \\
\textit{uttara}      & $+0.301$ \\
\textit{soma}        & $+0.277$ \\
\textit{asura}       & $+0.275$ \\
\textit{gu\d{n}a}    & $+0.251$ \\
\textit{\=atman}     & $+0.208$ \\
\textit{bh\d{r}tya}  & $+0.159$ \\
\textit{a\d{m}\'su}  & $+0.147$ \\
\textit{setu}        & $+0.072$ \\
\textit{\'s\=udra}   & $+0.057$ \\
\textit{varu\d{n}a}  & $+0.054$ \\
\textit{tejas}       & $+0.036$ \\
\textit{ari}         & $+0.030$ \\
\textit{vrata}       & $+0.014$ \\
\textit{p\=ada}      & $-0.100$ \\
\textit{go}          & $-0.132$ \\
\midrule
\textit{deva} (control) & $-0.159$ \\
\bottomrule
\end{tabular}
\caption{Directional anchor test on the primary model. $\Delta>0$ means the word moved in the philologically determined direction. 19/21 positive; sign test $p\approx0.00011$. The control \textit{deva} behaves as expected: with the exact same anchor set as \textit{asura}, it consolidates toward its original divine sense rather than the demonic.}
\label{tab:anchor}
\end{table}

\begin{table}[t]
\centering\small
\begin{tabularx}{\linewidth}{@{}lX@{}}
\toprule
Era & Top neighbors (truncated) \\
\midrule
Vedic & \textit{mahat} `great' (k=1); \textit{dev\=an\=am} `of the gods' (k=2); \textit{vi\d{s}\d{n}o\d{h}} `of Vi\d{s}\d{n}u' (k=6); \textit{s\=avitram} `of Savit\d{r} (sun deity)' (k=8) \\
Upani\d{s}adic & \textit{bhog\={a}ya} `for wordly enjoyment, consumption' (k=2); \textit{abhim\=anam} `ego, pride' (k=10); \textit{avidu\d{s}\=am} `of the ignorant' (k=11); \textit{s\=ukara} `hog, pig' (k=12) \\
Epic & \textit{d\=anava} `demon' (k=1); \textit{uraga} `serpent' (k=2); \textit{pi\'s\=aca} `flesh-eating ghouls' (k=4); \textit{daitya} `titan, enemy of god'(k=10) \\
\bottomrule
\end{tabularx}
\caption{Selected nearest neighbors from the $k{=}12$ list for \textit{asura} across the V $\rightarrow$ E transition. The neighbors shift from divine associations (Vedic), to ignorant and indulgent associations (Upani\d{s}adic), and finally to demonic associations (Epic).}
\label{tab:asura}
\end{table}

The two misses are informative. For \textit{go} `cow', the varied metaphorical extensions (`earth', `rays', `speech') reflect heavy polysemy stemming from Sanskrit poetic register. The trained model, ultimately, uses static embeddings, which assign \textit{go} a single fixed vector and cannot accurately capture the semantic variation as contextual embeddings do, as seen in Table~\ref{tab:go}. For \textit{p\=ada} `foot', the new-sense anchors are sparse in the relevant periods and the neighbor evidence in Table~\ref{tab:pada} is stronger than the anchor statistic. Together, these two failures highlight two limitations of the approach: representing polysemous meanings with a single representation and capturing non-linear semantic trajectories.

\begin{table}[t]
\centering
\small
\begin{tabularx}{\linewidth}{@{}lX@{}}
\toprule
Era & Top neighbors (truncated) \\
\midrule
Vedic & \textit{dhenava\d{h}} `milch-cows' (k=1); \textit{vatsam} `calf' (k=4); \textit{dh\={\i}taya\d{h}} `prayers' (k=8); \textit{sindhava\d{h}} `rivers' (k=12); \textit{v\=a\'sr\=a\d{h}} `lowing cows' (k=13) \\
Epic & \textit{v\d{r}\d{s}am} `bull' (k=2); \textit{prad\=ane} `offering' (k=7); \textit{prad\=at\=a} `donor', \textit{go\d{s}\d{t}he} `cow-pen' (k=10); \textit{k\d{s}\={\i}ra} `milk' (k=14) \\
S\=utra & \textit{viv\={\i}ta} `pasture ground' (k=3); \textit{carma} `hide, leather' (k=5); \textit{k\=a\d{s}\d{t}ha} `timber, wood' (k=8); \textit{m\d{r}ga} `deer, animal' (k=10); \textit{vajra} `diamond' (k=14) \\
\bottomrule
\end{tabularx}
\caption{Selected nearest neighbors from the $k=15$ list for \textit{go} by era. While the literal meaning `cow' remains stable across eras, the attested metaphorical senses (`earth', `rays', `speech') do not consistently appear. Vedic shows a flowing river association, while Epic and S\=utra emphasize the gift / commodity frames.}
\label{tab:go}
\end{table}

\begin{table}[t]
\centering
\small
\begin{tabularx}{\linewidth}{@{}lX@{}}
\toprule
Era & Top neighbors (truncated) \\
\midrule
Vedic & \textit{\'suna\d{h}} `dog' (k=1); \textit{caritv\=a} `having moved / walked' (k=7); \textit{mimate} `they measured' (k=9); \textit{dvip\=ada\d{h}} `bipedal' (k=14) \\
Upani\d{s}adic &  \textit{catu\d{s}kala\d{h}} `having four parts' (k=2), \textit{catu\d{s}p\=ad} `four-footed' (k=3), \textit{s\=urya\d{h}} `the sun' (k=9), \textit{candra\d{h}} `the moon' (k=11), \textit{p\=a\d{n}i} `hand' (k=13) \\
Epic &  \textit{m\=ule} `at the base' (k=2); \textit{cara\d{n}a} `foot' (k=8), \textit{ka\d{t}\={\i}} `hip, waist' (k=10); \textit{ja\.ngh\=a} `shank, lower leg' (k=11) \\
S\=utra & \textit{catu\d{h}} `four' (k=2); \textit{\'sva} `dog' (k=6); \textit{a\.ngula\d{h}} `finger, finger's breadth (unit)' (k=13) \\
\bottomrule
\end{tabularx}
\caption{Selected nearest neighbors from the $k=15$ list for \textit{p\=ada} by era. The Vedic vector is overly sparse ($n{=}12$), but manages to retrieve some sense. The `quarter' sense surfaces clearly in the Upani\d{s}adic and S\=utra eras, but the Epic vector reverts to the literal body-part frame, producing a non-monotonic trajectory that the anchor test cannot follow cleanly.}
\label{tab:pada}
\end{table}

\subsection{Embedding architecture}
\label{sec:rq2}

\begin{table}[t]
\centering\small
\begin{tabular}{@{}lrrr@{}}
\toprule
Model & Orth.\% & $\rho$ & Vocab \\
\midrule
FastText/raw (3--6)     & 68\% & 0.67 & 801 \\
word2vec/raw            & 7\%  & 0.70 & 801 \\
FastText/sandhi (3--6)  & 67\% & 0.74 & 1{,}345 \\
FastText/sandhi (6--12) & 19\% & 0.65 & 1{,}345 \\
\textbf{word2vec/sandhi} & \textbf{4\%} & \textbf{0.70} & \textbf{1{,}345} \\
word2vec/lemma          & 7\%  & 0.58 & 1{,}246 \\
\bottomrule
\end{tabular}
\caption{Orth.\% = share of top-10 neighbors that are merely orthographic (shared $\geq$4-character affix or edit distance $\leq2$), averaged over case words $\times$ periods; lower is better. $\rho$ = Spearman agreement between the second-order and Procrustes measures (reliability). Vocab = words trackable across all four periods.}
\label{tab:grid}
\end{table}

Table~\ref{tab:grid} demonstrates the following effects. \textbf{1) Preprocessing dictates coverage}: sandhi splitting raises the trackable vocabulary from 801 to 1{,}345 ($+68\%$), simply by breaking up fused variants of the same lexeme. \textbf{2) Architecture influences neighbor quality}: with standard $n$-grams, two-thirds of FastText's nearest neighbors are orthographic look-alikes rather than semantic associates. For Epic \textit{r\=aj\=a} `king', the FastText model returns inflectional echoes of the stem, while the word2vec one returns a more accurate neighbor set with `ruler' and `lord':
\begin{itemize}[noitemsep]
  \item \textbf{FastText (3-6):} \textit{r\=aj}, \textit{r\=aja\d{h}}, \textit{r\=aj\={\i}}, \textit{r\=aj\=ar\d{s}i\d{h}}
  \item \textbf{word2vec:} \textit{n\d{r}pati\d{h}}, \textit{sah\=am\=atya\d{h}}, \textit{mah\={\i}p\=ala\d{h}}
\end{itemize}

The failure is specifically a short-$n$-gram artifact; raising the range to $[6,12]$, where subwords approach morpheme length, cuts orthographic pollution from 67\% to 19\% and restores largely semantic neighbors. Even so, this improved FastText model remains ${\sim}5\times$ more polluted than word2vec (19\% vs.\ 4\%) and is the least reliable sandhi variant ($\rho=0.65$). Although FastText is better equipped for morphological and OOV handling \citep{sandhan-etal-2023-evaluating}, word2vec performs better for this diachronic semantic task.

Despite aggregating many inflected forms of the same root, lemmatization, which erases syntactic information but remains stable for semantic tasks, did not increase performance: as per Table~\ref{tab:grid}, word2vec/lemma is slightly more polluted and notably less reliable ($\rho=0.58$). The lemmatized corpus is built from a lookup table powered by the ByT5 lemmatizer, which may collapse forms that carry distinct senses and lead to the perceived embedding degradation.

Raising training to 30 epochs appears to overfit within-period neighbors, degrading the between-period signal measured. The cross-metric reliability (\S\ref{sec:change}) falls from $0.70$ to $0.25$ for the primary model, meaning diachronic shifts do not recover as cleanly. Moreover, the 30-epoch model yields only marginal improvements over the 10-epoch model on the within-period EvalSan intrinsic benchmarks (\S\ref{sec:evalsan}).

Lastly, the change scores correlate positively with frequency ($\rho=+0.30$), opposite to the law of conformity reported by \citet{hamilton2016}. Given the limited number of periods and substantial data imbalance, this pattern is more plausibly attributed to estimation noise than to a fully-supported linguistic effect. The recovery result is not explained by frequency alone, since the anchor-displacement metric is relative. $\Delta$ compares movement toward the new anchors against the old anchors, meaning general shifts in the embedding space from frequency imbalance would affect both sides similarly and largely cancel. A positive score therefore requires movement specifically toward the new sense rather than simply shifting in the space. 

\subsection{EvalSan}
\label{sec:evalsan}
For completeness, the Epic-period embeddings are evaluated on the EvalSan intrinsic benchmarks for the semantic categorization task \citep{sandhan-etal-2023-evaluating}. Semantic-categorization purity is 0.27 and 0.42 for the word2vec/sandhi and word2vec/lemma models, respectively. The latter matches the purity score of 0.41 that their word2vec embeddings, trained on a pooled 5.7M-token corpus, achieve. The remaining tasks are constrained primarily by vocabulary coverage rather than embedding quality, as EvalSan reports that roughly 55\% of relatedness test words are OOV. This issue is more pronounced for a single diachronic period, where the available vocabulary is substantially smaller (e.g., the synonym-MCQ benchmark has no in-vocabulary items in the Epic bin). Accordingly, the purity score rises to 0.38 (word2vec/sandhi) and to 0.61 (word2vec/lemma) when restricted to only in-vocabulary items, suggesting that greater corpus coverage would improve embeddings.

\section{Conclusion}

This paper investigates whether diachronic word-embeddings can recover semantic change in ancient Sanskrit. Using a validation set of semantic shifts, the proposed directional anchor evaluation correctly aligns with the attested trajectory for 19 of 21 words. These results show that, despite Sanskrit's extensive morphology, sandhi, compounding, and limited diachronic data, distributional methods can recover meaningful patterns of semantic change when adapted appropriately. I will release the four-period corpus, preprocessing pipeline, validation set, and evaluation code. Future work includes expanding and bolstering the corpora, comparing contextual and temporally aware embedding models against the current static approach, and extending the evaluation framework to other ancient languages.

\section*{Limitations}

While the proposed framework demonstrates that distributional methods can capture aspects of Sanskrit semantic change, several limitations remain.

First, the corpus remains limited in size, balance, and quality. The Epic bin is substantially larger than the other bins, while some periods rely on smaller or less securely dated textual collections. Furthermore, residual noise from commentary and editorial marks, digitization, and automated preprocessing may introduce artifacts into the learned representations. Expanding the corpus while increasing data verification would provide more robust semantic results. Second, this study evaluates static embedding models. While these models are effective in measuring aggregate movement, they ultimately collapse multiple senses into a single representation. Contextualized and temporally aware approaches may better capture polysemy, sense emergence, and non-monotonic change. Finally, the evaluation framework measures recovery of previously documented shifts rather than open-ended discovery. Despite providing an independent benchmark, this approach may favor well-studied changes and does not fully evaluate the ability to uncover lesser-studied or new shifts.

\bibliography{custom}

@inproceedings{hamilton2016,
    title = "Diachronic Word Embeddings Reveal Statistical Laws of Semantic Change",
    author = "Hamilton, William L.  and Leskovec, Jure  and Jurafsky, Dan",
    editor = "Erk, Katrin and Smith, Noah A.",
    booktitle = "Proceedings of the 54th Annual Meeting of the Association for Computational Linguistics (Volume 1: Long Papers)",
    month = aug,
    year = "2016",
    address = "Berlin, Germany",
    publisher = "Association for Computational Linguistics",
    url = "https://aclanthology.org/P16-1141/",
    doi = "10.18653/v1/P16-1141",
    pages = "1489--1501"
}

@article{bhargavaAsura,
 ISSN = {03781143},
 URL = {http://www.jstor.org/stable/41693045},
 author = {P. L. Bhargava},
 journal = {Annals of the Bhandarkar Oriental Research Institute},
 number = {1/4},
 pages = {119--128},
 publisher = {Bhandarkar Oriental Research Institute},
 title = {THE WORD ASURA IN THE {\d{R}}GVEDA},
 urldate = {2026-07-15},
 volume = {64},
 year = {1983}
}

@inproceedings{hamilton2016b,
    title = "Cultural Shift or Linguistic Drift? Comparing Two Computational Measures of Semantic Change",
    author = "Hamilton, William L.  and
      Leskovec, Jure  and
      Jurafsky, Dan",
    editor = "Su, Jian  and
      Duh, Kevin  and
      Carreras, Xavier",
    booktitle = "Proceedings of the 2016 Conference on Empirical Methods in Natural Language Processing",
    month = nov,
    year = "2016",
    address = "Austin, Texas",
    publisher = "Association for Computational Linguistics",
    url = "https://aclanthology.org/D16-1229/",
    doi = "10.18653/v1/D16-1229",
    pages = "2116--2121"
}

@article{bojanowski2017,
    title = "Enriching Word Vectors with Subword Information",
    author = "Bojanowski, Piotr  and
      Grave, Edouard  and
      Joulin, Armand  and
      Mikolov, Tomas",
    editor = "Lee, Lillian  and
      Johnson, Mark  and
      Toutanova, Kristina",
    journal = "Transactions of the Association for Computational Linguistics",
    volume = "5",
    year = "2017",
    address = "Cambridge, MA",
    publisher = "MIT Press",
    url = "https://aclanthology.org/Q17-1010/",
    doi = "10.1162/tacl_a_00051",
    pages = "135--146"
}

@inproceedings{kutuzov-etal-2018-diachronic,
    title = "Diachronic word embeddings and semantic shifts: a survey",
    author = "Kutuzov, Andrey  and
      {\O}vrelid, Lilja  and
      Szymanski, Terrence  and
      Velldal, Erik",
    editor = "Bender, Emily M.  and
      Derczynski, Leon  and
      Isabelle, Pierre",
    booktitle = "Proceedings of the 27th International Conference on Computational Linguistics",
    month = aug,
    year = "2018",
    address = "Santa Fe, New Mexico, USA",
    publisher = "Association for Computational Linguistics",
    url = "https://aclanthology.org/C18-1117/",
    pages = "1384--1397",
}

@inproceedings{kulkarni2015statistically,
author = {Kulkarni, Vivek and Al-Rfou, Rami and Perozzi, Bryan and Skiena, Steven},
title = {Statistically Significant Detection of Linguistic Change},
year = {2015},
isbn = {9781450334693},
publisher = {International World Wide Web Conferences Steering Committee},
address = {Republic and Canton of Geneva, CHE},
url = {https://doi.org/10.1145/2736277.2741627},
doi = {10.1145/2736277.2741627},
booktitle = {Proceedings of the 24th International Conference on World Wide Web},
pages = {625–635},
numpages = {11},
location = {Florence, Italy},
series = {WWW '15}
}

@inproceedings{schlechtweg-etal-2020-semeval,
    title = "{S}em{E}val-2020 Task 1: Unsupervised Lexical Semantic Change Detection",
    author = "Schlechtweg, Dominik  and
      McGillivray, Barbara  and
      Hengchen, Simon  and
      Dubossarsky, Haim  and
      Tahmasebi, Nina",
    editor = "Herbelot, Aurelie  and
      Zhu, Xiaodan  and
      Palmer, Alexis  and
      Schneider, Nathan  and
      May, Jonathan  and
      Shutova, Ekaterina",
    booktitle = "Proceedings of the Fourteenth Workshop on Semantic Evaluation",
    month = dec,
    year = "2020",
    address = "Barcelona (online)",
    publisher = "International Committee for Computational Linguistics",
    url = "https://aclanthology.org/2020.semeval-1.1/",
    doi = "10.18653/v1/2020.semeval-1.1",
    pages = "1--23",
}

@inproceedings{perrone-etal-2019-gasc,
    title = "{GASC}: Genre-Aware Semantic Change for {A}ncient {G}reek",
    author = "Perrone, Valerio  and
      Palma, Marco  and
      Hengchen, Simon  and
      Vatri, Alessandro  and
      Smith, Jim Q.  and
      McGillivray, Barbara",
    editor = "Tahmasebi, Nina  and
      Borin, Lars  and
      Jatowt, Adam  and
      Xu, Yang",
    booktitle = "Proceedings of the 1st International Workshop on Computational Approaches to Historical Language Change",
    month = aug,
    year = "2019",
    address = "Florence, Italy",
    publisher = "Association for Computational Linguistics",
    url = "https://aclanthology.org/W19-4707/",
    doi = "10.18653/v1/W19-4707",
    pages = "56--66",
}

@article{goyal2016design,
  title={Design and Analysis of a Lean Interface for Sanskrit Corpus Annotation},
  author={Goyal, Pawan and Huet, Gérard},
  journal={Journal of Language Modelling},
  volume={4},
  number={2},
  pages={145--182},
  year={2016},
  month={October},
  doi={10.15398/jlm.v4i2.108},
  url={https://jlm.ipipan.waw.pl/index.php/JLM/article/view/108},
}

@inproceedings{aralikatte-etal-2018-sanskrit,
    title = "{S}anskrit Sandhi Splitting using seq2(seq)2",
    author = "Aralikatte, Rahul  and
      Gantayat, Neelamadhav  and
      Panwar, Naveen  and
      Sankaran, Anush  and
      Mani, Senthil",
    editor = "Riloff, Ellen  and
      Chiang, David  and
      Hockenmaier, Julia  and
      Tsujii, Jun{'}ichi",
    booktitle = "Proceedings of the 2018 Conference on Empirical Methods in Natural Language Processing",
    month = oct # "-" # nov,
    year = "2018",
    address = "Brussels, Belgium",
    publisher = "Association for Computational Linguistics",
    url = "https://aclanthology.org/D18-1530/",
    doi = "10.18653/v1/D18-1530",
    pages = "4909--4914",
}

@online{dcs,
  title = {DCS - The Digital Corpus of Sanskrit},
  author = {Hellwig, Oliver},
  year = {2010--2021},
  url = {http://www.sanskrit-linguistics.org/dcs/index.php}
}

@article{rodda2017panta,
  author  = {Rodda, Martina A. and Senaldi, Marco S. G. and Lenci, Alessandro},
  title   = {Panta rei: Tracking Semantic Change with Distributional Semantics in Ancient Greek},
  journal = {Italian Journal of Computational Linguistics},
  volume  = {3},
  number  = {1},
  pages   = {11--24},
  year    = {2017},
  doi     = {10.4000/ijcol.421}
}

@book{tahmasebi2021survey,
editor = {Tahmasebi, Nina and Borin, Lars and Jatowt, Adam and Xu, Yang and Hengchen, Simon},
title = {{Computational} approaches to semantic change},
year = {2021},
series = {Language Variation},
number = {6},
address = {Berlin},
publisher = {Language Science Press},
doi = {10.5281/zenodo.5040241}
}

@inproceedings{kim-etal-2014-temporal,
    title = "Temporal Analysis of Language through Neural Language Models",
    author = "Kim, Yoon  and
      Chiu, Yi-I  and
      Hanaki, Kentaro  and
      Hegde, Darshan  and
      Petrov, Slav",
    editor = "Danescu-Niculescu-Mizil, Cristian  and
      Eisenstein, Jacob  and
      McKeown, Kathleen  and
      Smith, Noah A.",
    booktitle = "Proceedings of the {ACL} 2014 Workshop on Language Technologies and Computational Social Science",
    month = jun,
    year = "2014",
    address = "Baltimore, MD, USA",
    publisher = "Association for Computational Linguistics",
    url = "https://aclanthology.org/W14-2517/",
    doi = "10.3115/v1/W14-2517",
    pages = "61--65"
}

@inproceedings{gulordava-baroni-2011-distributional,
    title = "A distributional similarity approach to the detection of semantic change in the {G}oogle {B}ooks Ngram corpus.",
    author = "Gulordava, Kristina  and
      Baroni, Marco",
    editor = "Pado, Sebastian  and
      Peirsman, Yves",
    booktitle = "Proceedings of the {GEMS} 2011 Workshop on {GE}ometrical Models of Natural Language Semantics",
    month = jul,
    year = "2011",
    address = "Edinburgh, UK",
    publisher = "Association for Computational Linguistics",
    url = "https://aclanthology.org/W11-2508/",
    pages = "67--71"
}

@book{bloomfield1933,
  author    = {Leonard Bloomfield},
  title     = {Language},
  year      = {1933},
  publisher = {Compton Printing Works Ltd}
}

@book{breal1897,  
  title = "Essai de s{\'e}mantique", 
  author = "Br{\'e}al, Michel",  
  publisher = "Hachette", 
  year = "1897", 
}

@article{Witzel2001Autochthonous,
  author       = {Michael Witzel},
  title        = {Autochthonous Aryans? The Evidence from Old Indian and Iranian Texts},
  journal      = {Electronic Journal of Vedic Studies},
  year         = {2001},
  volume       = {7},
  number       = {3},
  pages        = {1--93},
  doi          = {10.11588/ejvs.2001.3.830},
  url          = {https://doi.org/10.11588/ejvs.2001.3.830}
}

@inproceedings{mikolov-etal-2013-distributed,
    title = "Distributed Representations of Words and Phrases and their Compositionality",
    author = "Mikolov, Tomas and Sutskever, Ilya and Chen, Kai and Corrado, Greg S. and Dean, Jeffrey",
    booktitle = "Advances in Neural Information Processing Systems",
    year = "2013",
    pages = "3111--3119",
}

@misc{hariharan2025,
      title={Transformer-Enabled Diachronic Analysis of Vedic Sanskrit: Neural Methods for Quantifying Types of Language Change}, 
      author={Ananth Hariharan and David Mortensen},
      year={2025},
      eprint={2512.05364},
      archivePrefix={arXiv},
      primaryClass={cs.CL},
      url={https://arxiv.org/abs/2512.05364}, 
}

@article{pradeep202,
author = {Pradeep, Anagha and Mamidi, Radhika},
title = {Sandar\'{s}ana: A Survey on Sanskrit Computational Linguistics and Digital Infrastructure for Sanskrit},
year = {2025},
issue_date = {October 2025},
publisher = {Association for Computing Machinery},
address = {New York, NY, USA},
volume = {57},
number = {10},
issn = {0360-0300},
url = {https://doi.org/10.1145/3729530},
doi = {10.1145/3729530},
journal = {ACM Computing Surveys},
month = may,
articleno = {254},
numpages = {38},
}

@book{kamboj,
  title={Semantic Change in Sanskrit},
  author={Kamboj, Jiyalal},
  publisher={Vidyanidhi Prakashan},
  edition={Second Revised and Enlarged},
  year={2017},
  isbn={978-93-85539-26-8}
}

@inproceedings{N_J_2025,
   title={LEVOS: Leveraging Vocabulary Overlap with Sanskrit to Generate Technical Lexicons in Indian Languages},
   url={http://dx.doi.org/10.18653/v1/2025.bea-1.20},
   DOI={10.18653/v1/2025.bea-1.20},
   booktitle={Proceedings of the 20th Workshop on Innovative Use of NLP for Building Educational Applications (BEA 2025)},
   publisher={Association for Computational Linguistics},
   author={N J, Karthika and Bhatt, Krishnakant and Ramakrishnan, Ganesh and Jyothi, Preethi},
   year={2025},
   pages={258–265} }

@inproceedings{sandhan-etal-2023-sanskritshala,
    title = "{S}anskrit{S}hala: A Neural {S}anskrit {NLP} Toolkit with Web-Based Interface for Pedagogical and Annotation Purposes",
    author = "Sandhan, Jivnesh  and
      Agarwal, Anshul  and
      Behera, Laxmidhar  and
      Sandhan, Tushar  and
      Goyal, Pawan",
    editor = "Bollegala, Danushka  and
      Huang, Ruihong  and
      Ritter, Alan",
    booktitle = "Proceedings of the 61st Annual Meeting of the Association for Computational Linguistics (Volume 3: System Demonstrations)",
    month = jul,
    year = "2023",
    address = "Toronto, Canada",
    publisher = "Association for Computational Linguistics",
    url = "https://aclanthology.org/2023.acl-demo.10/",
    doi = "10.18653/v1/2023.acl-demo.10",
    pages = "103--112",
}

@book{whitney1879sanskrit,
  title     = {A Sanskrit Grammar},
  author    = {Whitney, William Dwight},
  year      = {1879},
  publisher = {Breitkopf and Härtel},
  address   = {Leipzig}
}

@article{riceGunna,
 DOI = {10.2307/408769},
 author = {Carlton C. Rice},
 journal = {Language},
 number = {1},
 pages = {36-40},
 publisher = {Linguistic Society of America},
 title = {The Etymology of Sanskrit guṇá},
 volume = {6},
 year = {1930}
}

@article{paoloTejas,
 ISSN = {00840084, 17283124},
 URL = {http://www.jstor.org/stable/24010814},
 author = {Paolo Magnone},
 journal = {Wiener Zeitschrift für die Kunde Südasiens / Vienna Journal of South Asian Studies},
 pages = {137--147},
 publisher = {Austrian Academy of Sciences Press},
title = {THE DEVELOPMENT OF "TEJAS" FROM THE VEDAS TO THE PUR{\=A}{\d{N}}AS},
 urldate = {2026-07-12},
 volume = {36},
 year = {1992}
}

@article{lubinVrata,
 ISSN = {00030279},
 URL = {http://www.jstor.org/stable/606499},
 author = {Timothy Lubin},
 journal = {Journal of the American Oriental Society},
 number = {4},
 pages = {565--579},
 publisher = {American Oriental Society},
 title = {Vratá Divine and Human in the Early Veda},
 volume = {121},
 year = {2001}
}

@misc{nehrdich2024modelneedbyt5sanskritunified,
      title={One Model is All You Need: ByT5-Sanskrit, a Unified Model for Sanskrit NLP Tasks}, 
      author={Sebastian Nehrdich and Oliver Hellwig and Kurt Keutzer},
      year={2024},
      eprint={2409.13920},
      archivePrefix={arXiv},
      primaryClass={cs.CL},
      url={https://arxiv.org/abs/2409.13920}, 
}

@inproceedings{sandhan-etal-2023-evaluating,
    title = "Evaluating Neural Word Embeddings for {S}anskrit",
    author = "Sandhan, Jivnesh  and
      Paranjay, Om Adideva  and
      Digumarthi, Komal  and
      Behra, Laxmidhar  and
      Goyal, Pawan",
    editor = "Kulkarni, Amba  and
      Hellwig, Oliver",
    booktitle = "Proceedings of the Computational {S}anskrit {\&} Digital Humanities: Selected papers presented at the 18th World {S}anskrit Conference",
    month = jan,
    year = "2023",
    address = "Canberra, Australia (Online mode)",
    publisher = "Association for Computational Linguistics",
    url = "https://aclanthology.org/2023.wsc-csdh.2/",
    pages = "21--37"
}

@inproceedings{hellwig-nehrdich-2018-sanskrit,
    title = "{S}anskrit Word Segmentation Using Character-level Recurrent and Convolutional Neural Networks",
    author = "Hellwig, Oliver  and
      Nehrdich, Sebastian",
    editor = "Riloff, Ellen  and
      Chiang, David  and
      Hockenmaier, Julia  and
      Tsujii, Jun{'}ichi",
    booktitle = "Proceedings of the 2018 Conference on Empirical Methods in Natural Language Processing",
    month = oct # "-" # nov,
    year = "2018",
    address = "Brussels, Belgium",
    publisher = "Association for Computational Linguistics",
    url = "https://aclanthology.org/D18-1295/",
    doi = "10.18653/v1/D18-1295",
    pages = "2754--2763",
}

\appendix

\section{Hyperparameters}
\label{app:hyper}

Models were trained with skip-gram and negative sampling, dimension 300, window 5, minimum count 5, 10 negative samples, subsampling $10^{-4}$, 10 epochs, seed 42. FastText additionally uses character $n$-grams in the stated ranges. Procrustes rotations are fit on the 767 shared words with frequency $\geq10$ in every period. Neighbor size $k=25$ for the second-order measure.

\section{Validation set}
\label{app:valset}
\begin{table*}[t]
\centering
\small
\renewcommand{\arraystretch}{1.5}
\begin{tabularx}{\textwidth}{@{}l X X l p{2cm} p{1.2cm}@{}}
\toprule
Word & Before meaning & After meaning & Transition & Change type & Source \\
\midrule
\textit{asura}        & ``lord, mighty one; life-giving god'' (epithet even of Indra, Varu\d{n}a, Agni in early RV) & ``demon, hostile being'' & V $\rightarrow$ E & pejoration & \citep{kamboj}\\
\textit{\=atman}        & ``breath'' (RV); ``body, life, vital essence'' & ``the Soul, Self'' & V $\rightarrow$ U & extension (metaphorical, metonymic) & \citep{kamboj} \\
\textit{brahman}        & ``growth of the soul'' (RV); ``pious utterance / devotion, prayer, Veda'' & ``one supreme, absolute World-Soul''& V $\rightarrow$ U & abstraction (metaphorical) & \citep{kamboj} \\
\textit{varu\d{n}a}        & ``omnipotent sovereign, superior to all deities'', ``moral governor of all'' & ``god of waters / the ocean'' & V $\rightarrow$ E & specialization & \citep{kamboj} \\
\textit{go}        & ``cow'', bovine products like ``milk'', ``ghee'' or ``leather'' & ``earth'', ``sun, rays of light'', ``cloud'', ``senses'' & V $\rightarrow$ E & extension (metaphorical, metonymic) & \citep{kamboj} \\
\textit{hari}        & ``tawny, yellow, brown'', ``golden'' (color), ``soma'' (elixir, plant) &  ``horse, monkey'', ``rays, sun'', divine epithet (Indra/Vi\d{s}\d{n}u)'' & V $\rightarrow$ E &  extension (metaphorical, metonymic) & \citep{kamboj} \\
\textit{yoga}        & ``the act of yoking, joining'', ``union'' & ``act of meditation, self concentration'', ``spiritual discipline'' & U $\rightarrow$ E &  extension (metaphorical) & \citep{kamboj} \\
\textit{gu\d{n}a}        & ``bovine'', ``sinew'', ``bowstring'', ``strand, cord (of rope)'' & ``quality, virtue'' & U $\rightarrow$ E &  extension (metaphorical) & \citep{riceGunna} \\
\textit{soma}        &  ``soma plant, pressed juice'' & permitted substitute plants, identified with the moon & V $\rightarrow$ E &  extension & \citep{kamboj} \\
\textit{k\d{s}atra}        &  ``might, dominion, rule'', ``a noble man'' & ``the warrior/ruling class, nobility'' & V $\rightarrow$ E &  specialization & \citep{kamboj} \\
\textit{\'s\=udra}        &  section of aboriginals, a non-Aryan tribe & ``the servile/labor class'' & V $\rightarrow$ E &  extension & \citep{kamboj} \\
\textit{ari}        &  ``miser, one who does not give'' & ``enemy'' \textbf{and} ``master, lord, pious man'' & V $\rightarrow$ E &  bifurcation, pejoration & \citep{kamboj} \\
\textit{uttara}        &  ``upper, higher, above'' & ``later, subsequent'', ``superior'', ``northern''  & V $\rightarrow$ E &  extension & \citep{kamboj} \\
\textit{uttama}        &  ``highest, upper-most'' (superlative) & ``best, most excellent, last''  & V $\rightarrow$ E &  extension (metaphorical) & \citep{kamboj} \\
\textit{p\=ada}        &  ``foot of man, animals'' & ``quarter, fourth part'', ``foot of a mountain'', ``ray''  & V $\rightarrow$ E &  extension (metaphorical) & \citep{kamboj} \\
\textit{tejas}      &  ``sharpness, edge'', ``fire, flame, brilliance'' & ``splendour, glory'', ``vital energy, spiritual power'', ``majesty''  & V $\rightarrow$ E & abstraction & \citep{paoloTejas} \\
\textit{setu}        &  ``bond, fetter'' & ``causeway, dam, bridge'', ``boundary, protection''  & U $\rightarrow$ S &  extension (metaphorical, metonymic) & \citep{kamboj} \\
\textit{bh\d{r}tya}        &  ``one to be supported/maintained, a dependent in a family'' & ``servant, slave''  & E $\rightarrow$ S &  pejoration & \citep{kamboj} \\
\textit{arka}        &  ``ray, flash'', ``song/hymn of praise'' & ``the sun'', the arka plant & V $\rightarrow$ E &  borrowing & \citep{kamboj} \\
\textit{a\d{m}\'su}        &  ``filament/stalk of the soma plant'', ``soma juice'' & ``ray of the moon / sun'' & V $\rightarrow$ E &  extension (metaphorical) & \citep{kamboj} \\
% \textit{preta}        &  ``departed, deceased one'' & ``ghost, spirit of the dead, hungry ghost'' & V $\rightarrow$ E &  extension (metaphorical) & \citep{kamboj} \\
\textit{vrata}        &  ``divine ordinance, law, command of the gods'', ``sacred rite'' & ``(self-imposed) vow, religious observance, ascetic regimen'' & V $\rightarrow$ E &  specialization & \citep{lubinVrata} \\

\bottomrule
\end{tabularx}
\caption{Validation set of documented semantic shifts. (V=Vedic, U=Upani\d{s}adic, E=Epic, S=S\=utra).}
\label{tab:validation}
\end{table*}

\end{document}